\pdfoutput=1

\documentclass[11pt]{article}

\usepackage[final]{acl}

\usepackage{times}
\usepackage{latexsym}

\usepackage[T1]{fontenc}

\usepackage[utf8]{inputenc}

\usepackage{microtype}

\usepackage{inconsolata}

\usepackage{graphicx}

\usepackage{algorithm}
\usepackage{algorithmic}
\usepackage{multirow}
\usepackage{colortbl}
\usepackage{amsmath}

\title{Kill two birds with one stone: generalized and robust AI-generated text detection via dynamic perturbations}

\author{
 \textbf{Yinghan Zhou},
 \textbf{Juan Wen\textsuperscript{*}},
 \textbf{Wanli Peng\textsuperscript{*}},
 \textbf{Yiming Xue},
\\
 \textbf{Ziwei Zhang},
 \textbf{Zhengxian Wu}
\\
 China Agricultural University,
\\
 \small{
   {\{zhouyh, wenjuan, wlpeng, xueym, zzwei, wzxian\}@cau.edu.cn}
 }
}

\begin{document}
\maketitle
\begin{abstract}
The growing popularity of large language models has raised concerns regarding the potential to misuse AI-generated text (AIGT). 
It becomes increasingly critical to establish an excellent AIGT detection method with high generalization and robustness.
However, existing methods either focus on model generalization or concentrate on robustness.
The unified mechanism,  to simultaneously address the challenges of generalization and robustness, is less explored. 
In this paper, we argue that robustness can be view as a specific form of domain shift, and empirically reveal an intrinsic mechanism for model generalization of AIGT detection task.
Then, we proposed a novel AIGT detection method (DP-Net) via dynamic perturbations introduced by a reinforcement learning with elaborated reward and action.
Experimentally, extensive results show that 
the proposed DP-Net significantly outperforms some state-of-the-art AIGT detection methods for generalization capacity in three cross-domain scenarios.
Meanwhile, the DP-Net achieves best robustness under two text adversarial attacks. 
The code is publicly available at \href{https://github.com/CAU-ISS-Lab/AIGT-Detection-Evade-Detection/tree/main/DP-Net}{DP-Net}.
\end{abstract}

\section{Introduction}

Recently, the emergence of large language models (LLMs) has significantly enhanced the capabilities of  natural language generation. 
With continuous improvements in model parameters, data scale, and AI-human alignment techniques, these LLMs are now capable of generating text that is grammatically correct, semantically coherent, and very similar to human-written text, making it difficult for humans to distinguish between machine-generated and human-written content. 
As powerful tools for streamlining content creation, LLMs are widely used across various domains, including journalism, academia, and social media. 
However, the threats posed by AI-generated text (AIGT), such as academic dishonesty \cite{wu-etal-2023-llmdet,Zeng_Sha_Li_Yang_Gašević_Chen_2024}, fake news \cite{su-etal-2024-adapting,Hu_Sheng_Cao_Shi_Li_Wang_Qi_2024}, and false comments \cite{mireshghallah-etal-2024-smaller} have raised significant concerns. 
To prevent the LLM abuse with malicious purpose, numerous AIGT detection methods have been proposed, utilizing the specific feature differences between human-written text and AI-generated text.

Mainstream AIGT detection methods typically fall into one of two categories: white-box or black-box detection, depending on whether the generator's parameters are accessible. 
While some white-box methods \cite{gehrmann-etal-2019-gltr,wang-etal-2023-seqxgpt,yang2024dnagpt,2023arXiv230414072L} achieve high-precision detection by extracting textual features from the generator's output logits, black-box AIGT detection methods \cite{10386674,soto2024fewshot,tian2024multiscale} are gaining increasing attention for their ability to operate without accessing the model's parameters, making them particularly relevant in the context of many closed-source commercial LLMs.

Current black-box AIGT detection methods learn deep features on large annotated datasets, preforming well when training and testing samples are independent and identically distributed (i.i.d). 
However, the detection of out-of-distribution (OOD) samples is inevitable in real-world application scenarios. 
Therefore, some researchers are concentrated on enhancing generalization capacity for cross-domain detection \cite{liu-etal-2023-coco, bhattacharjee-etal-2023-conda}. 
These methods aim to enhance generalization in unknown target domains through different feature alignment strategy. 
Some studies \cite{detectgpt,zhu-etal-2023-beat} introduce zero-shot learning methods to achieve domain-generalization AIGT detection in unpredictable target distributions by extracting effective domain-invariant features.

To evade detection, users may attempt to modify the generated text when using AI tools. However, current AIGT detectors face challenges in detecting minor adversarial perturbations \cite{zhou-etal-2024-humanizing-machine,krishna2023paraphrasing}. Therefore, some research has explored robust AIGT detection methods \shortcite{NEURIPS2023_30e15e59,koike2024outfoxllmgeneratedessaydetection}.
Despite the good generalization ability of existing AIGT methods, they are vulnerable for text adversarial attacks. 
Meanwhile, some robust detection method can not achieve good generalization for OOD data.
An interesting and challenging problem is how to construct a comprehensive framework that exhibits high generalization and robustness against text adversarial attack.

In this paper, we argue that model generalization and robustness are intrinsically linked. 
Both concepts can be viewed as responses to variations introduced in the source domain. 
Specifically, robustness is often validated by introducing minor perturbations within the source domain, while OOD data can be understood as changes in the distribution resulting from more substantial and specific disturbances.
The analysis spontaneously induce an evident thinking: \textit{Can adding elaborated perturbations simultaneously improve generalization and robustness of AIGT detection network?}

In order to exploring the rationality of the above thinking, we first empirically reveal an intrinsic mechanism for model generalization through making quantitative and qualitative experiments.
Then, in light of the intrinsic mechanism, we propose a novel AIGT detection method via dynamic perturbations (called DP-Net). The crux of DP-Net is that dynamic perturbations, added into embedding matrix in training phase, are yielded by a reinforcement learning with elaborated reward and action.
Extensive experiments demonstrate that the proposed DP-Net can significantly improve the generalization and robustness for AIGT detection task.
Our contributions are as follows:
\begin{itemize}
    \item We empirically reveal an intrinsic mechanism for generalization capacity of AIGT detection. In other words, adding slight perturbations into source domain can effectively simulate the domain shift between source and target domains.
    \item We propose a novel AIGT detection method (DP-Net) via dynamic perturbations which are adaptively yielded by the reinforcement learning with elaborated reward and action.
    \item Extensive experiments demonstrate that the proposed DP-Net achieves state-of-the-art generalization in three cross-domain scenarios and achieves best robustness under two text adversarial attacks.
\end{itemize}

\section{Related Work}

\subsection{Supervised Detector}
Supervised classifiers are trained on large labeled datasets to extract representations that can effectively distinguish between two classes of text samples.
For instance,  \citet{zhong-etal-2020-neural} combined the actual structure of text with a classifier based on RoBERTa \cite{liu2020roberta} to improve detection accuracy. 
Tian et al. \citet{tian2024multiscale} introduce a length-sensitive Multiscale Positive-Unlabeled Loss, which enhances the detection performance of short texts while maintaining the detection efficacy for long texts. These methods achieve strong detection performance in detecting datasets belonging to the same domain as the training set, but usually fail when faced with datasets that are not in the domain of the training set.
To enhance the model's generalization ability in known target domain, \citet{liu-etal-2023-coco} combined maximum mean discrepancy with contrastive learning to obtain domain-invariant representations, facilitating adaptation of classifiers from source to target generators. \citet{verma-etal-2024-ghostbuster} proposed Ghostbuster, a method that processes documents through a series of weaker language models, conducts a structured search over possible combinations of their features, and then trains a classifier on the selected features to predict whether the documents are AI-generated.

\subsection{Zero-shot Detector}
With the widespread use of LLMs, AIGT detectors often encounter unknown domain detection in practical applications, which has sparked a growing interest in zero-shot domain generalization detectors.  
\citet{detectgpt} proposed a text perturbation method to measure the log probabilities difference between original and perturbed texts.
\citet{su-etal-2023-detectllm} proposed a zero-shot method that measures the log-probability difference between original text and perturbed text using text perturbation techniques, significantly improving AIGT detection performance.
\citet{venkatraman-etal-2024-gpt} argue that humans tend to evenly distribute information during language production, whereas AI-generated text may lack this uniformity. Therefore, they introduce uniform information density features to quantify the smoothness of token distribution, aiding in the identification of AI-generated text.

\begin{figure*}[t]
    \centering
    \includegraphics[width=0.8\linewidth]{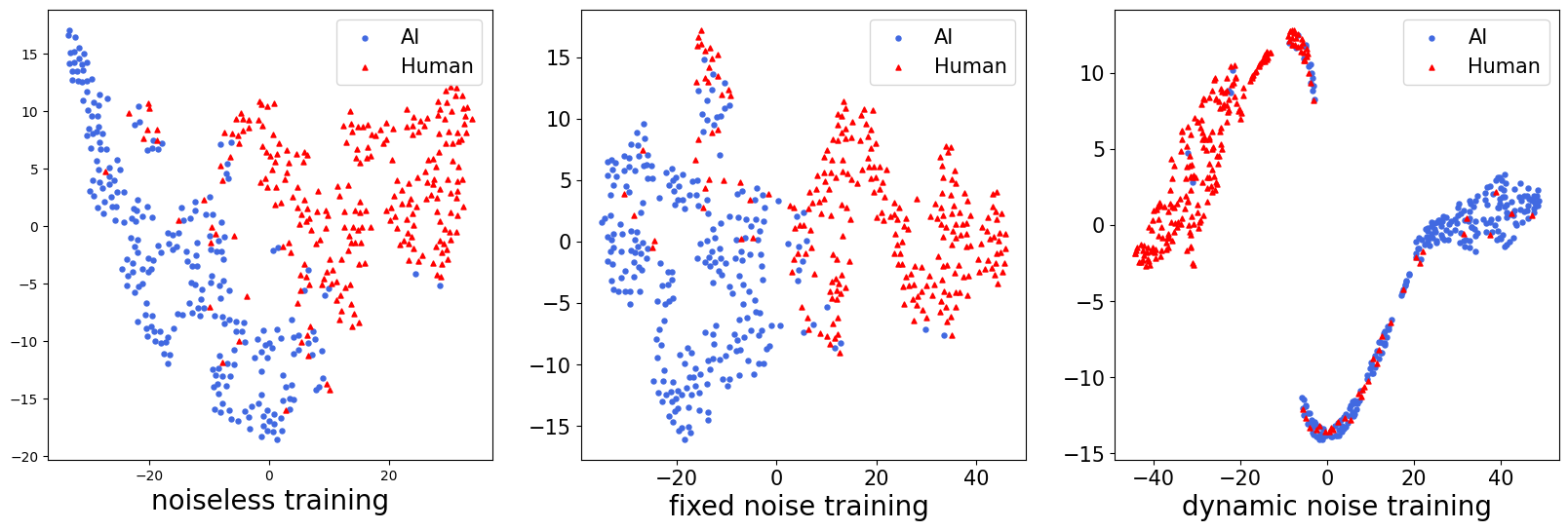}
    \caption{Visualization results of features extracted by detectors with different training strategies.}
    \label{fig:adding_noise}
\end{figure*} 

\begin{figure}[t]
    \centering
    \includegraphics[width=0.8\linewidth]{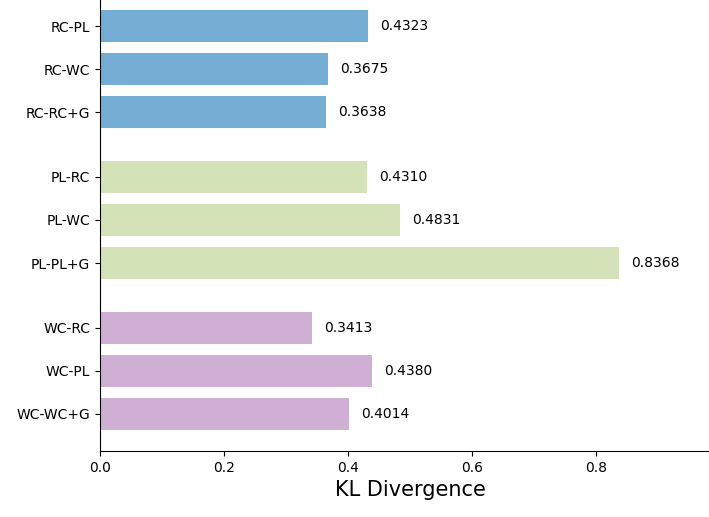}
    \caption{KL divergence between different distributions.X-Y represents the distance between the two distributions X and Y, while "RC", "PL", and "WC" refer to Reddit ChatGPT, PeerRead Llama, and WikiHow ChatGPT, respectively. Additionally, "+G" indicates the
addition of Gaussian noise.}
    \label{fig:KL_D}
\end{figure}


\section{Methodology}
\subsection{Intrinsic Mechanism Analysis}
Following \citet{pedro2000unified}, the machine learning error can be decomposed into three components: noise, bias, and variance. Consider an example $x$ with the true label $y$, and a learner $f$ that predicts $f(x)$ given a training set $D$. For certain loss functions $L$, the following decomposition of $\mathit{E}_{D,y} [L(f(x), y)]$ holds:

\begin{equation}
    \begin{aligned}
        \mathit{E}_{D,y} \big[ L(f(x), y) \big] 
        &= c_1 E_{\text{noise}} 
        + \underbrace{L \big( \mathit{E}(y), \mathit{E}_{D}(f(x)) \big)}_{\text{Bias}} \\
        &\quad + \underbrace{c_2 \mathit{E}_{D} \big[ L \big( \mathit{E}_{D}(f(x)), f(x) \big) \big]}_{\text{Variance}}
    \end{aligned}
\end{equation}

Empirically, both generalization and robustness are influenced by the sensitivity of model to fluctuations in the training data.
We find that appropriately calibrated noise perturbations can significantly reduce the variance component of the generalization error without substantially affecting the bias or noise components.
This reduction in variance leads to improved generalization performance across diverse data distributions.
Specifically, by introducing noise, the variance component of the error becomes directly proportional to both the noise variance and the squared gradient norm of the model.
This insight forms the foundation of our design. If the noise variance is too large, the model may become overly insensitive to the true patterns in the data, increasing variance and leading to poor generalization.
Conversely, if the noise variance is too small, the model may overfit the specific characteristics of the training data, failing to generalize well to unseen data, which also increases variance.
Therefore, we employ reinforcement learning during training to dynamically adjust the noise distribution, enhancing both the its robustness and ability to generalize to unseen data.

\begin{figure}
    \centering
    \includegraphics[width=0.9\linewidth]{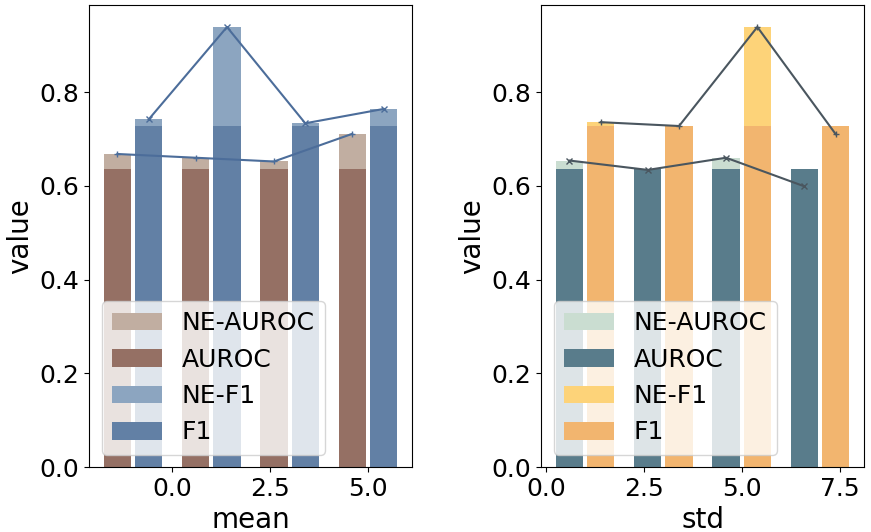}
    \caption{The impact of adding noise on the generalization performance of model.
    'NE' stands for training with noise-enhanced samples.}
    \label{fig:change}
\end{figure}

\begin{figure*}[t]
    \centering
    \includegraphics[width=0.8\linewidth]{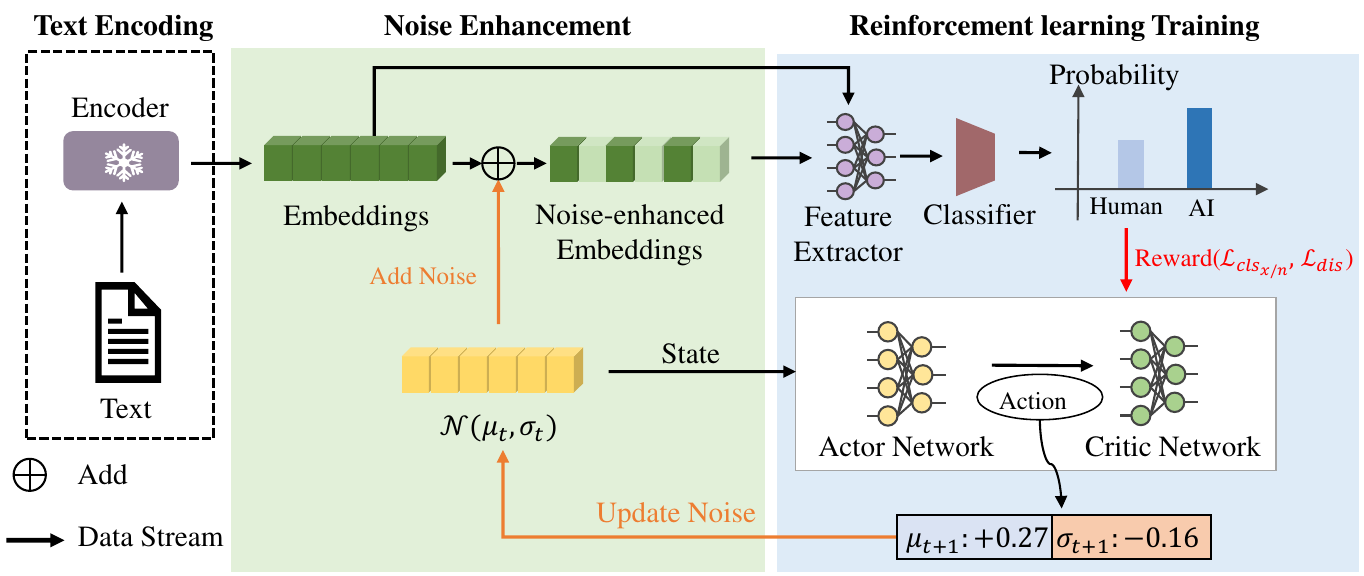}
    \caption{The overall framework of DP-Net. This framework includes three modules: text encoding, noise enhancement, and reinforcement learning training.}
    \label{fig:framework}
\end{figure*}
To evaluate the rationality of the intrinsic mechanism, i.e., adding slight perturbations into source domain can effectively simulate the domain shift between source and target domains, we make some quantitative and qualitative experiments.
For the quantitative experiment, we leverage the Roberta network to extract semantic spaces of different domains, and then Kullback-Leibler Divergence (KLD) is used to measure domain shift of the different semantic spaces. 
As shown in Figure \ref{fig:KL_D}, we can find that the KLD of ``RC-RC+G'' is similar to those of other two cross-domain scenarios, i.e. ``RC-WC'' and ``RC-PL'' from the blue lines.
The trend are also exhibited by the experimental results from the pink lines. 
While, the results shown by green lines have obvious KLD gap, which is different from other two.
The experimental results demonstrate that adding slight perturbations to the source domain effectively simulates the domain shift between different domains.
In addition, since different cross-domain scenarios have different domain shift, the network can learn excellent domain-invariant features, adding dynamic perturbations in the training phase.

For the qualitative experiment, we validate the impact of noise on the detection performance of the model in unknown domains, and the experimental results are shown in Figure \ref{fig:change}. 
The experiments demonstrate that introducing dynamic perturbations during the training process can effectively improve model generalization.
Additionally, a single type of noise cannot guarantee optimal performance of the model in unknown target domain. 
We also obtain similar results by visualizing the features extracted by detectors using different training strategies, as shown in Figure \ref{fig:change}. 
It is found that a single type of noise cannot guarantee an improvement in the detector's performance in unknown domains. 
Therefore, we continuously adjust the noise distribution during training through reinforcement learning to encourage the model to learn more generalized feature representations. 
As shown in Figure \ref{fig:change}, by introducing a reinforcement learning joint training strategy, the model is able to achieve clearer decision boundaries in unknown target domains.
We treat attacks on text as minor perturbations to the source domain, which represents a specific case of domain shift. Therefore, introducing dynamic perturbations can also inherently enhance the robustness of model.

According to the above quantitative and qualitative analysis, we proposed a novel AIGT detection method via dynamic perturbations (DP-Net) mainly consists of three modules: Text Encoding, Noise Enhancement and Reinforcement Learning Training. 
The detail architectures of the DP-Net are introduced as following sections.

\subsection{Text Encoding and Noise Enhancement}
 We introduce noise-enhanced samples during training to improve the generalization and robustness of the detector. The overall process is as follows. 

First, text \(x\) consisting of \(n\) tokens is encoded by a pre-trained RoBERTa \cite{liu2020roberta} with fixed parameters to obtain text embedding  \(\mathbf{E}_{x}\) in text encoding module. 

Next, we randomly generate a Gaussian noise \(\mathcal{N}(\mu, \sigma^2)\) with its mean \(\mu\) and variance \(\sigma^2\) within a specified range. Then the noise is added to the embedding of input text \(\mathbf{E}_{x}\) to obtain the noise-enhanced embedding \(\mathbf{E}_{n}\):
\begin{equation}
    \mathbf{E}_{n} = \mathbf{E}_{x} + \mathcal{N}(\mu, \sigma^2)
\end{equation}

After that, the original embedding and the noise-enhanced embedding are input into a feature extractor to obtain low-dimensional classification representations \(\mathbf{z}_{x}\) and \(\mathbf{z}_{n}\), respectively:
\begin{equation}
    \mathbf{z}_{x / n} = Extractor(\mathbf{E}_{x / n})
\label{encode}
\end{equation}
Where $\mathbf{z}_{x / n}$ reperents \(\mathbf{z}_{x}\) or \(\mathbf{z}_{n}\) and $\mathbf{E}_{x / n}$ represents \(\mathbf{E}_{x}\) or \(\mathbf{E}_{n}\).

To improve the ability of detector to align noisy samples with original samples, we introduce a distance loss \( L_{dis} \) to minimize the distance between the two embeddings:
\begin{equation}
    L_{dis} = \|\mathbf{z}_x - \mathbf{z}_{n}\|_2^2
\end{equation}

Then \(\mathbf{z}_{x}\) and \(\mathbf{z}_{n}\) are processed by a $\mathit{softmax}$ function to obtain the predicted probabilities $\mathbf{\hat{y}}_x$ and $\mathbf{\hat{y}}_n$, respectively.  After that, two cross-entropy losses $L_{cls_x}$ and $L_{cls_n}$ are incorporated to increase prediction accuracy.
\begin{equation}\small
    L_{cls_{x/n}} = -\left[\mathbf{y} \log(\mathbf{\hat{y}}_{x/n}) + (1 - \mathbf{y} ) \log(1 - \mathbf{\hat{y}}_{x/n}) \right]
\end{equation}
where \( \mathbf{y} \) is the true label of text $x$.
The final model loss \(L\) is defined as:
\begin{equation}
    L = \lambda_1 \cdot L_{cls_x} + \lambda_2 \cdot L_{cls_n} +  \lambda_3 \cdot L_{dis}
\label{loss}
\end{equation}
where \(\lambda_1\), \(\lambda_2\) and \(\lambda_3\) are adjustable hyper-parameters.

\subsection{Reinforcement Learning Training}
To enhance sample diversity and guide the model in finding more suitable noise parameters for the unknown target domain, we introduce reinforcement learning. Inspired by Liu et al. \shortcite{liu2023on}, we treat the mean and variance of Gaussian noise during training as a continuous control task. We employ an actor-critic algorithm called Deep Deterministic Policy Gradient (DDPG) \cite{DDPG}, which leverages deep neural networks to learn deterministic policies in continuous action spaces to control the noise distribution, aiming to maximize cumulative rewards through a combination of off-policy learning and target network stabilization. 

We design a reward function based on the training loss of original samples and noise-enhanced samples to guide this process. The reward function is defined as:
\begin{equation}\small
reward(L)=\left\{
\begin{array}{lll}
-log(L-\epsilon) & & {L-\epsilon > 0}\\
e^{\epsilon-L} & & {L-\epsilon \leq 0}\\
\end{array} \right.
\label{reward}
\end{equation}
where \(\epsilon\) is an adjustable threshold hyper-parameter.

We store the tuple $(s, a, r, s')$ into the experience replay buffer $\mathcal{R}$, where $s$ represents the current noise distribution state, $a$ is the action taken, $s'$ is the subsequent state resulting from the action, and $r$ is the reward received. Next, DDPG is used to optimize the policy by maximizing the Q-function, aiming to maximize the long-term cumulative reward for actions taken in all states.

For each sampled tuple, the target Q-value is computed using the target networks:
\begin{equation}
    y_i = r_i + \gamma Q'(s_i', \mu'(s_i'|\theta^{\nu'})|\theta^{Q'})
\end{equation}
where \(\gamma\) is the discount factor, \(Q'\) and \(\nu'\) are the target Q and policy networks, respectively. The parameters \(\theta^Q\) of the Q-network are then updated by minimizing the loss function $L(\theta^Q)$:
\begin{equation}
    L(\theta^Q) = E_{(s,a,r,s')}[(Q(s_i, a_i|\theta^Q) - y_i)^2]
\end{equation}

The policy network $\nu$ is updated using the deterministic policy gradient:
\begin{small}
\begin{equation}
    \nabla_{\theta^\nu} J(\theta^\nu) = E_{s \sim \mathcal{R}} \left[ \nabla_a Q(s, a \mid \theta^Q)\bigg|_{a=\nu(s)} \\
    \nabla_{\theta^\nu} \mu(s \mid \theta^\nu) \right]
\end{equation}
\end{small}
This update seeks to maximize the expected return by adjusting the parameters \(\theta^{\nu}\) of the policy network.

The target networks' parameters are updated using a soft update strategy:
\begin{equation}
   \theta^{Q'} \leftarrow \tau \theta^Q + (1 - \tau) \theta^{Q'}
\end{equation}
\begin{equation}
   \theta^{\nu'} \leftarrow \tau \theta^{\nu} + (1 - \tau) \theta^{\nu'}
\end{equation}
where \(\tau \ll 1\) is a small parameter that ensures slow updates to the target networks, thereby providing stability to the training process.

We illustrate the general process in pseudo Algorithm \ref{alg}.
\begin{algorithm}[t]\small
\caption{DP-Net Algorithm}
\label{alg}
\textbf{Input}: source domain training data \(D=\{X,Y\}\)\\
\textbf{Parameter}: Maximum number of iterations \(max\_episode\), exploration steps per episode \(max\_step\), loss weight \(\lambda_1, \lambda_2,\lambda_3\), reward threshold \(\epsilon\)\\
\textbf{Output}:Trained detector \(Detector\)
\begin{algorithmic}[1] 
\STATE $env$ $\leftarrow$ $Env()$ \# Initialize environment
\STATE $agent$ $\leftarrow$ $DDPG()$ \# Initialize reinforcement learning agent
\FOR{$episode$ in \(max\_episode\)}
\STATE Initial current state \(s\)
\FOR{$step$ in \(max\_step\)}
\STATE Retrieve the current action $a$ based on \(s\)
\STATE Act based on the $s$ to obtain the next state \(s'\).
\FOR{all $x$ sampled from $D$}
\STATE Predict all $x$ based on \(s'\) to obtain the loss $L$ with Eq.\ref{loss} and the reward \(r\) with Eq.\ref{reward}
\STATE Update encoder $Detector$ based on $L$
\ENDFOR
\STATE Store $s$, $a$, $r$ and $s'$ in experience replay buffer \(\mathcal{R}\)
\STATE  update politic network \(\nu\) and critic network \(Q\)
\STATE $s$ $\leftarrow$ $s'$ \# update state
\ENDFOR
\ENDFOR
\STATE \textbf{return} Trained detector \(Detector\)
\end{algorithmic}
\end{algorithm}

\section{Experiments}
\subsection{Dataset}
We use the M4 \cite{wang-etal-2024-m4} as the benchmark dataset to construct various cross-domain scenarios. M4 is a large-scale, multi-domain corpus generated by three LLMs for the AIGT detection task. We utilize a subset of datasets from M4, with detailed information provided in Table \ref{dataset}. 
\begin{table}[t]\small
\centering
\setlength{\tabcolsep}{0.5mm}
\begin{tabular}{ccccc}
\hline
    Corpus  & Human & Davinci003 & ChatGPT & FlanT5 \\
\hline
    Wikipedia & 3,000 & 3,000 & 2995 & 0 \\
    Reddit & 3,000 & 3,000 & 3,000 & 3,000 \\
    WikiHow & 3,000 & 3,000 & 3,000 & 0 \\
    peerread & 5,798 & 2,344 & 2,344 & 0 \\
    arXiv & 3,000 & 3,000 & 3,000 & 3,000 \\
\hline
\end{tabular}
\caption{Sample numbers from different sources in M4.}
\label{dataset}
\end{table}

In the following experiments, we denote the data sources by combining the corpus name with the generator name.  For example, 'Arxiv ChatGPT' signifies that the corpus consists of Arxiv abstracts \cite{arxiv_org_submitters_2024}, and the AI-text generator used for this corpus is ChatGPT \cite{ouyang2022traininglanguagemodelsfollow}.

\subsection{Baseline Methods}
We compared five benchmark methods for black-box AIGT detection:
(1) \textbf{Naive Classifer} is a supervised method, which consist of a fixed RoBERTa model with an additional MLP classification head. The training process is consistent with Roberta.
(2) \textbf{Roberta} \cite{liu2020roberta} is a supervised method. In this work, we fine-tune the parameters on the labeled training set and perform detection in the target domain.
(3) \textbf{SCRN} \cite{huang2024aigeneratedtextdetectorsrobust} employs a reconstruction network to add and remove noise from text, extracting a semantic representation that is robust to local perturbations. It uses two random noise sources to enhance the representation and learns through twin networks. In contrast, our DP-Net approach uses reinforcement learning to continuously modify the noise distribution during training.
(4) \textbf{GLTR} \cite{gehrmann-etal-2019-gltr} proposes three simple tests to assess whether the text is generated in a specific assumed manner. In this work, we use the most powerful Test-2 feature, which is the absolute rank of a word, consistent with \citet{guo2023closechatgpthumanexperts}.
(5) \textbf{Fast-DetectGPT} \cite{bao2024fastdetectgpt} is an optimized zero-shot detector, which utilize conditional probability curvature to elucidate discrepancies in word choices between LLMs and humans within a given context. In this study, both the sampling model and the scoring model used in the method are GPT-2 \cite{radford2019language}.

\subsection{Experimental Setting}
During the training process, the agent continuously adjusts the noise distribution based on the reward at each step and optimizes both the actor and critic networks. 
Simultaneously, the detector is optimized based on the model's final loss. The optimizers for all networks in the model are Adam, with a learning rate of \(3e^{-4}\) for the actor and critic networks and \(8e^{-5}\) for the encoder. 
A learning rate decay strategy is introduced to gradually reduce the learning rate, promoting model convergence. 
The model is trained for 300 epochs on the source domain data, which is Arxiv ChatGPT, and tested on different unseen target domains in this experiment. 
The weight parameters used during training are \(\lambda_1 = 0.5\), \(\lambda_2 = 0.5\), \(\lambda_2 = 0.01\) and the threshold \(\epsilon = 1\). 
In the experiments involving reinforcement learning, the reported results are averaged across five random seeds. The GPU used for the experiments is an NVIDIA RTX 4060. 
\begin{table*}[t]
    \centering
    \setlength{\tabcolsep}{0.6mm}
    \begin{tabular}{c|cccccccccc|cccc}
    \hline
        \multirow {2}{*}{Method} & \multicolumn{2}{c}{Naive} &
        \multicolumn{2}{c}{\multirow {2}{*}{Roberta}} & 
        \multicolumn{2}{c}{\multirow {2}{*}{SCRN}} & 
        \multicolumn{2}{c}{\multirow {2}{*}{GLTR}} & 
        \multicolumn{2}{c|}{Fast} & \multicolumn{4}{c}{DP-Net} \\[0.2pt]
         & \multicolumn{2}{c}{Classifier} & & & & & & &\multicolumn{2}{c|}{detectGPT} & \multicolumn{2}{c}{+U(ours)} & \multicolumn{2}{c}{+G(ours)} \\
    \hline
        Target &  \multirow {2}{*}{Acc} & \multirow {2}{*}{F1} & \multirow {2}{*}{Acc} & \multirow {2}{*}{F1} & \multirow {2}{*}{Acc} & \multirow {2}{*}{F1} & \multirow {2}{*}{Acc} & \multirow {2}{*}{F1} & \multirow {2}{*}{Acc} & \multirow {2}{*}{F1} & \multirow {2}{*}{Acc} & \multirow {2}{*}{F1} & \multirow {2}{*}{Acc} & \multirow {2}{*}{F1} \\[0.2pt]
        Domain & & & & & & & & & & & & & & \\
    \hline
        Wikihow-C & 64.00 & 
        72.89 & 54.90 & 
        68.35 & 51.10 & 44.67 & 44.30 & 55.48 & \textbf{79.90} & \textbf{79.08} & 68.52 & 61.19 & 66.62 & 57.34\\
        Wikipedia-C & 64.80 & 73.96 & 53.90 & 68.45 & 53.90 & 
        49.58 & 70.00 & 76.45 & 95.40 & 95.59 & \textbf{96.88} & \textbf{96.97} & 96.04 & 96.18\\
        Reddit-C & 63.10 & 73.05 & 51.90 & 67.51 & 44.92 & 45.98 & 54.90 & 68.62 & \textbf{91.90} & \textbf{91.91} & 88.10 & 88.41 & 89.62 & 90.17\\
        Arxiv-D & 94.70 & 94.10 & 98.00 & 97.96 & 51.50 & 45.37 & 78.10 & 76.87 & 36.00 & 35.48 & 90.94 & 89.89 & 89.52 & 88.27 \\
        Arxiv-F &98.40 & 98.37 & \textbf{99.80} & \textbf{99.80} & 51.04 & 63.72 & 83.30 & 83.18 & 82.50 & 80.83 & 91.66 & 90.82 & 90.84 & 89.91\\
        Peerread-D& 60.62 & 64.83 & 47.28 & 57.93 & 49.80 & 45.98 & 46.64 & 54.70 & \textbf{97.12} & 97.17 & 76.15 & 75.35 & 77.62 & 75.93\\
        Reddit-D & 63.10 & 73.05 & 51.90 & 67.52 & 51.00 & 44.60 & 55.00 & 68.71 & 82.00 & 81.93 & 86.10 & 86.14 & \textbf{87.72} & \textbf{88.04} \\ 
    \hline
    \rowcolor{gray!20}
        Average & 72.67 & 78.61 & 65.38 & 75.36 & 51.04 & 48.56 & 61.75 & 69.14 & 80.55 & 80.28 & 85.48 & 84.11 & \textbf{86.10} & \textbf{83.69}\\
    \hline
    \end{tabular}
    \caption{Comparison of Models across Different Domains. "+U" indicates adding uniform distribution noise, and "+G" indicates adding Gaussian noise. "-C","-D","-F" indicate that AIGT is generated by ChatGPT, Davinci, and FlanT5, respectively.}
    \label{generalization}
\end{table*}

\begin{table*}[t]
\centering
\setlength{\tabcolsep}{0.2mm}
\begin{tabular}{c|cccccccc|cccc}
\hline
\multirow{2}{*}{Method} & \multicolumn{2}{c}{\multirow {2}{*}{Roberta}} & \multicolumn{2}{c}{\multirow {2}{*}{SCRN}} & \multicolumn{2}{c}{\multirow {2}{*}{GLTR}} & \multicolumn{2}{c|}{Fast} & \multicolumn{4}{c}{DP-Net} \\[0.2pt]
 & & & & & & &\multicolumn{2}{c|}{detectGPT} & \multicolumn{2}{c}{+U(ours)} & \multicolumn{2}{c}{+G(ours)} \\
\hline
Attack Method & \multicolumn{12}{c}{Synonym replacement (ratio=0.2)} \\
\hline
Target Domain & AUROC & F1 & AUROC & F1 & AUROC & F1 & AUROC & F1 & AUROC & F1 & AUROC & F1 \\
\hline
 Wikipedia-C & 52.70 & 67.89 & 49.58 & 44.92 & 79.30 & 76.66 & 90.39 & 82.72 & 91.58 & 91.22 & \textbf{94.60} & \textbf{94.61} \\ 
 Reddit-C & 90.00 & 89.29 & 49.55 & 45.90 & 64.39 & 50.38 & 83.01 & 76.30 & 83.39 & 62.02 & \textbf{90.32} & \textbf{89.51}\\
 Reddit-D & 79.10 & 82.21 & 48.88 & 44.82 & 63.57 & 67.11 & 68.96 & 64.72 & 83.44 & 79.83 & \textbf{88.84} & \textbf{87.74}\\
 \rowcolor{gray!20}
 Average  & 73.93 & 79.80 & 49.34 & 45.21 & 69.09 & 64.72 & 80.79 & 74.58 & 86.98 & 84.81 & \textbf{91.25} & \textbf{90.62}\\
\hline
Attack Method & \multicolumn{12}{c}{Paraphrase}\\ 
\hline
Method & AUROC & F1 & AUROC & F1 & AUROC & F1 & AUROC & F1 & AUROC & F1 & AUROC & F1  \\
\hline
 Wikipedia-C & 53.90 & 68.45 & 49.88 & 45.16 & 56.40 & 66.84 & 61.76 & 64.35 & 75.50 & 69.62 & \textbf{76.50} & \textbf{71.93} \\ 
 Reddit-C & 64.90 & 65.82 & 49.20 & 45.37 & 62.61 & \textbf{66.71} & 31.32 & 35.26 & \textbf{68.44} & 62.02 & 67.92 & 62.02\\
 Reddit-D & 51.70 & \textbf{67.39} & 48.63 & 44.44 & 58.81 & 66.71 & 23.01 & 28.09 & 61.58 & 50.54 & \textbf{61.76} & 51.37\\
 \rowcolor{gray!20}
 Average  & 56.83 & \textbf{67.22} & 49.24 & 44.99 & 59.27 & 66.75 & 38.70 & 42.57 & 68.51 & 60.73 & \textbf{68.73} & 61.77\\
\hline
\end{tabular}
\caption{Evaluation of cross-domain adversarial robustness. The source domain is Arxiv ChatGPT. The optimal results are indicated in bold.}
\label{robustness}
\end{table*}

\subsection{Domain Generalization Results}
We test the generalization ability of our proposed DP-Net on seven target domains, which include three different cross-domain scenarios: same generator across different corpora, same corpus across different generators, and different corpora across different generators. 

\begin{figure}[t]
\centering
\includegraphics[width=0.9\columnwidth]{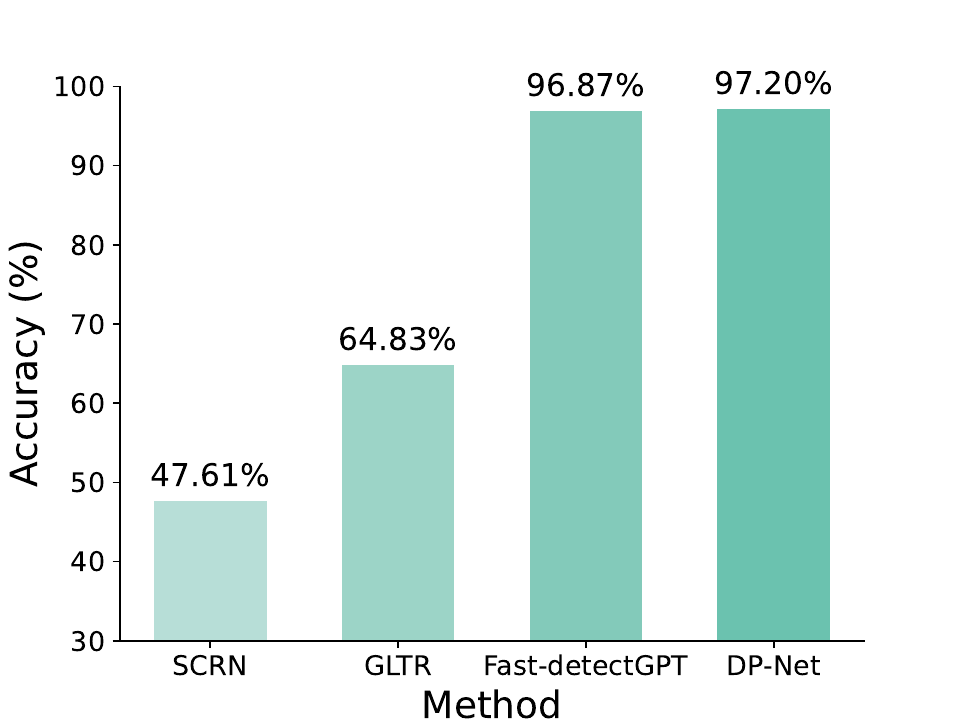}
\caption{Comparasion of Models on text generated by LLaMa-70B-chat.}
\label{llama}
\end{figure}

From the Table \ref{generalization}, DP-Net achieves an average detection accuracy of 86.10\% across seven unseen target domains, surpassing the other baseline methods by at least 5.55\% and demonstrating state-of-the-art domain generalization performance.
This is because zero-shot detectors like Fast-detectGPT heavily rely on the selection of proxy models, and different proxy models can significantly impact model performance.
However, supervised detectors based on training data, such as Roberta, suffer from the lack of target domain data. Training solely on source domain data prevents the model from extracting effective domain-invariant features, resulting in poorer detection performance on unseen domains.
Additionally, we find that among the misclassified samples, a larger proportion of human-written texts were mistakenly classified as AI-generated texts. We believe that compared to direct training, adding noise increases the distributional variance within the same class in the training samples. This is particularly challenging for human-written text, which has greater inherent variation, as the noise makes it harder for the model to learn a unified representation.
Furthermore, we introduce the RAID dataset to evaluate the performance of DP-Net on more advanced models. Specifically, we sample 1,000 entries generated by the llama-70B-chat and 1,000 human entries to from RAID. As shown in Figure \ref{llama}, DP-Net achieve a detection accuracy of 97.20\%, surpassing other baseline methods.

\begin{figure}[t]
\centering
\includegraphics[width=1\columnwidth]{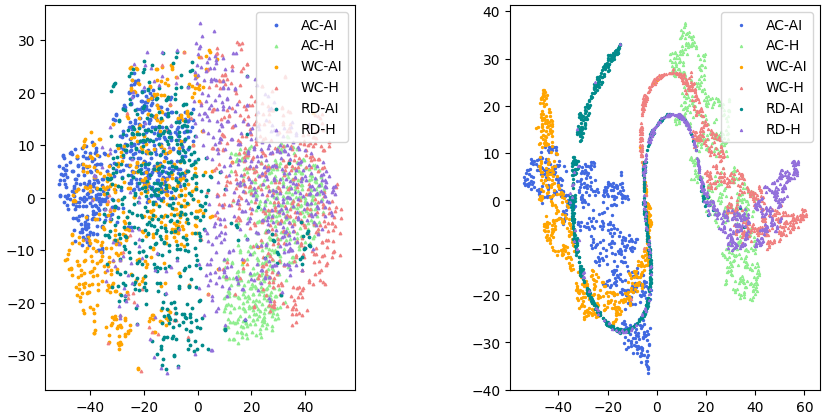}
\caption{Visualization of Domain Generalization Detection Results. The left figure represents the distribution of the original texts, while the right figure shows the distribution of the texts obtained through the trained encoder.}
\label{tsne}
\end{figure}

As seen from the visualization results in Figure \ref{tsne}, DP-Net effectively pulls together texts of the same class from different domains while pushing apart texts from different classes. This indicates that DP-Net has the generalization ability to perform detection in new, unseen domains.
Additionally, we find that among the misclassified samples, a larger proportion of human-written texts were mistakenly classified as AI-generated texts. We believe this is because that AIGT exhibits smaller intra-class variance compared to human-written text, making feature learning simpler. As a result, the model tends to classify text as AI-generated.

\subsection{Cross-Domain Adversarial Robustness}

We further evaluate the model's robustness under adversarial attacks in cross-domain scenarios. We perform synonym replacement with a replacement ratio of 0.2 and paraphrase attacks on the test data from the unknown target domain. In the paraphrasing attack, we use Flan-T5-base \cite{flant5} to rewrite AI-generated text. Additionally, we also explore the impact of using chatGPT for paraphrasing attacks on model performance, results can be found in the Appendix \ref{sec:para}.

Based on Table \ref{robustness}, our proposed method demonstrates superior robustness against attacks. It achieves the highest cross-domain detection accuracy under both types of attacks compared to other baseline methods. Specifically, under the synonym substitution attack, our method achieves an average detection accuracy of 91.25\% by introducing Gaussian noise, surpassing other baseline models. For the paraphrase attack, which significantly alters text distribution, our method attains an average cross-domain detection accuracy of 68.73\%, representing a 9.46\% improvement over the second-best baseline model. Additionally, We find that, DP-Net tends to classify text as AI-generated on certain out-of-domain datasets, which results in its F1 score being inferior to that of RoBERTa in the context of paraphrase attacks. We believe that compared to human-written text, AIGT exhibits smaller intra-class variance, making feature learning simpler. As a result, the model tends to classify text as AI-generated, leading to a higher recall but lower precision, which ultimately affects the F1 score. 

\subsection{Ablation Study}
\begin{figure}[t]
\centering
\includegraphics[width=0.8\columnwidth]{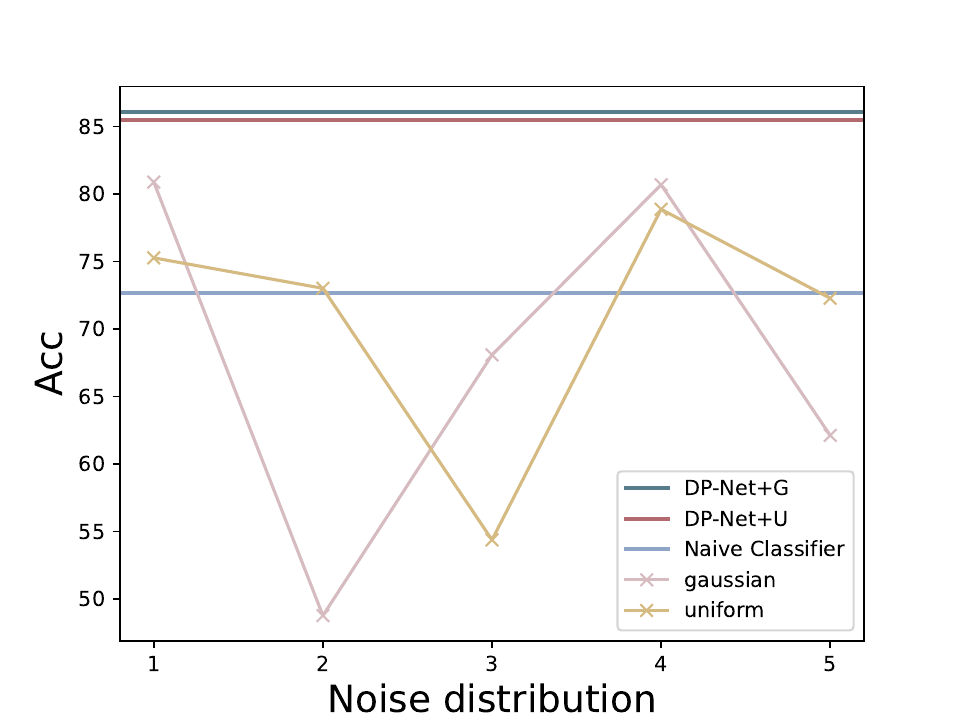}
\caption{The impact of different noise distribution types on model generalization.  $\times$ represents the average detection accuracy across 7 unkown domains when the model is trained with noise-augmented samples without using RL.
}
\label{noise}
\end{figure}

To examine the impact of noise types on model generalization, we compared the average detection accuracy across seven unseen domains under different noise settings, both with and without Joint Reinforcement Learning Training Strategy. The results shown in Figure \ref{noise} indicate that the model's generalization performance varies significantly across different noise settings. By introducing reinforcement learning, the detection accuracy of the model trained with noise-enhanced samples improved significantly. Additionally, as shown in Table \ref{generalization} and Table \ref{robustness}, the model trained with Gaussian noise shows better generalization and robustness against attacks compared to uniform distribution noise. We believe this is because Gaussian noise causes significant changes in the original sample distribution, thereby increasing the diversity of the training sample distribution. This, in turn, enhances the model's generalization and robustness after reinforcement training.

Next, we demonstrate the effectiveness of the DP-Net structure by adding Gaussian noise through two variants:
(1) \textbf{DP-Net+G$_{-rl}$}: Only noise-enhanced training is performed.
(2) \textbf{DP-Net+G$_{-encoder}$}: 
In this approach, the encoder parameters are fixed and used as the environment at first. Reinforcement learning is employed to find the optimal distribution, after which the encoder parameters are trained on the source domain data based on the optimal distribution.

\begin{table*}[t]
\centering
\setlength{\tabcolsep}{0.6mm}
\begin{tabular}{c|cccccccc|cc}
\hline
    \multirow{2}{*}{Target Domain} & \multicolumn{2}{c}{Wikipedia} & \multicolumn{2}{c}{Reddit} & \multicolumn{2}{c}{Arxiv} & \multicolumn{2}{c|}{Peerread} & \multicolumn{2}{|c}{\multirow {2}{*}{Average}} \\
     & \multicolumn{2}{c}{ChatGPT} & \multicolumn{2}{c}{ChatGPT} & \multicolumn{2}{c}{Davinci} & \multicolumn{2}{c|}{Davinci} & \\
\hline
 & AUROC & F1  & AUROC & F1 & AUROC & F1 & AUROC & F1 & AUROC & F1\\
\hline
    DP-Net+G (ours) & \textbf{96.04} & \textbf{96.18} & \textbf{89.62} & \textbf{90.17} & 89.52 & \textbf{88.27} & \textbf{82.36} & \textbf{75.39} & \textbf{89.39} & \textbf{87.50}\\
    DP-Net+G$_{-rl}$ & 67.50 & 75.47 & 66.00 & 74.63 & \textbf{94.00} & 93.62 & 70.23 & 65.58 & 74.43 & 77.33\\
    DP-Net+G$_{-encoder}$ & 71.00 & 77.52 & 75.00 & 80.00 & 88.50 & 87.01 & 73.79 & 68.49 & 77.07 & 78.26\\
\hline
\end{tabular}
\caption{Ablation Experiment Results. The source domain is Arxiv ChatGPT. The optimal results are indicated in bold.}
\label{ablation}
\end{table*}

We trained DP-Net and its variants on data from Arxiv ChatGPT, with the results shown in Table \ref{ablation}.
We find that directly adding noise to train the classifier results in a relatively low average cross-domain detection accuracy. This may be due to the significant differences between domains, making it difficult for a single noise sample to handle detection in various domain shift scenarios. For instance, in our experiments, DP-Net+G$_{-rl}$ achieves the highest detection accuracy on the test data sampled from Arxiv Davinci but performs poorly in other domains. Additionally, the two-step method is also affected by a single noise distribution. Although the optimal noise for the current environment is obtained through reinforcement learning initially, the model's generalization remains lower than that of DP-Net due to the continuous changes in the training environment and the fixed noise distribution.

We also explore the data efficiency of DP-Net.  We use 100\% and 75\% of the training data to train the model, and compare it with naive classifier, which is the DP-Net variant after removing noisy enhanced samples and RL, also trained with only 75\% of the training data. The results indicate that using only 75\% of the training data still achieves results comparable to those obtained using the entire training set. In contrast, reducing the training samples significantly degrades the performance of the Naive Classifier, demonstrating that our proposed DP-Net has high data efficiency. Specific experimental results can be found in Appendix \ref{sec:dataefficiency}.

Additionally, we compare DP-Net with a different RL method, Deep Q-Network (DQN). In this approach, we treat the changes of mean and variance in noise distribution as a discrete control task while keeping all other settings the same as the DDPG method. The experiments show that DQN, with its discretized actions, exhibits more stable rewards after multiple iterations than DDPG. The results demonstrate that the generalization performance of both methods on the six unseen domains is comparable. Specific experimental results can be found in Appendix \ref{sec:RL_methods}.

\section{Conclusion}
 We have proposed a novel AIGT detection method via dynamic perturbations (DP-Net) to simultaneously improve the model generalization and robustness against text adversarial attack.
 The DP-Net combines RL with a noise-enhanced training strategy. It adaptively introduces noise-enhanced samples to train the encoder, enhancing model generalization and robustness. Experiments show that the DP-Net performs excellently in robustness tests, maintaining high average detection accuracy under attacks in cross-domain situations. By continuously adjusting the distribution through RL, the introduction of Gaussian noise enables the model to achieve better generalization and robustness compared to uniform noise.

 \section{Limitation}
The revealed intrinsic mechanism dose not be proved by explicit mathematical expression, which unable to accurately guide the perturbation design, hindering the performance improvement.
In addition, for the case of multiple source domains, the proposed DP-Net has not comprehensively explored.

\section{Acknowledgements}
This work was supported in part by the National Natural Science Foundation of China under Grant 62272463, 62402117.
\bibliography{main}

\appendix

\section{Paraphrase Attack Using ChatGPT}
\label{sec:para}
\renewcommand{\thetable}{A\arabic{table}}
\setcounter{table}{0}
To further evaluate the robustness of DP-Net under adversarial attacks in cross-domain scenarios, we conduct further paraphrasing of the text using ChatGPT, and the results are presented in the table \ref{paraphrase}.

\begin{table}[t]
\centering
\setlength{\tabcolsep}{0.2mm}
\begin{tabular}{c|cc|cc}
\hline
\multirow{2}{*}{Method} & \multirow {2}{*}{GLTR} & Fast & \multicolumn{2}{c}{DP-Net} \\[0.2pt]
  & &detectGPT & +U(ours) & +G(ours) \\
\hline
 Wikipedia-C & 75.54  & 87.33  & 95.90  & \textbf{96.40} \\ 
 Reddit-C  & 43.44 & 62.22  & 90.10 &  \textbf{91.90} \\
 Reddit-D  & 39.86  & 57.70 &  90.10 & \textbf{91.80} \\
 \rowcolor{gray!20}
 Average & 52.95  & 69.08  & 92.03  & \textbf{93.37} \\
\hline
\end{tabular}
\caption{AUROC of cross-domain adversarial robustnes under paraphrase attack using chatGPT. The source domain is Arxiv ChatGPT. }
\label{paraphrase}
\end{table}

\begin{figure}[h]
\centering
\includegraphics[width=1\columnwidth]{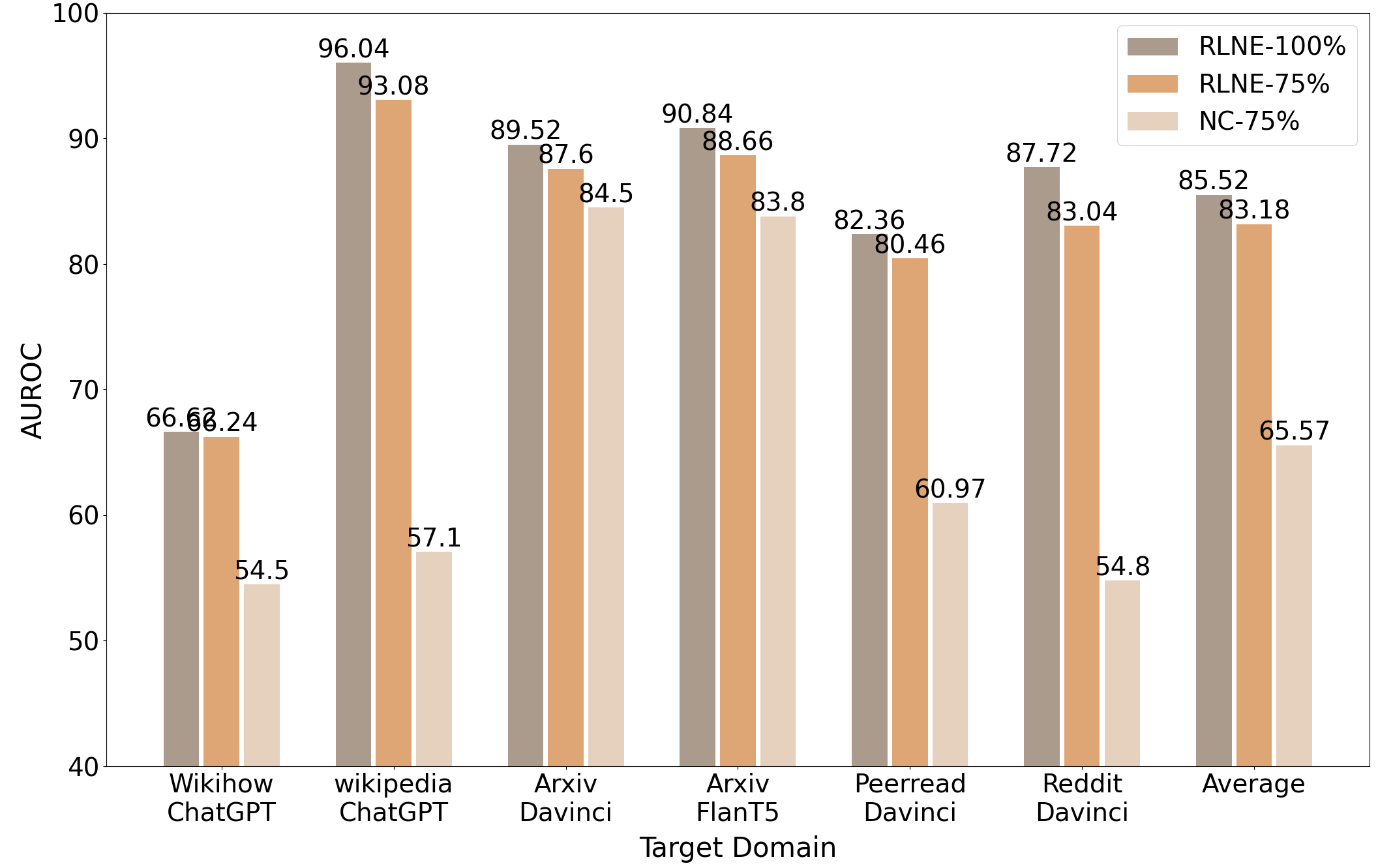}
\caption{Data efficiency experiment. NC-75\% indicates using 75\% of the training set to train the Naive classifier. The source domian is Arxic ChatGPT.}
\label{ablation_data}
\end{figure}
\section{Data Efficiency}
\label{sec:dataefficiency}
We trained the model using both 100\% and 75\% of the training data and compared its performance to that of a naive classifier. The naive classifier is also trained using only 75\% of the data. The experimental results are shown in the figure \ref{ablation_data}. In our DP-Net implementation for this experiment, we introduce Gaussian noise.
The results indicate that using only 75\% of the training data still achieves results comparable to those obtained using the entire training set, with only a 2.34\% decrease in performance. In contrast, reducing the training samples significantly degrades the performance of the naive classifier, which drops from 85.52\% with the full training set to 65.57\%.  This result indicates that our proposed DP-Net demonstrates high data efficiency.

\section{Comparison of Different Reinforcement Learning Methods.}
\label{sec:RL_methods}

\begin{figure}
\centering
\includegraphics[width=1\columnwidth]{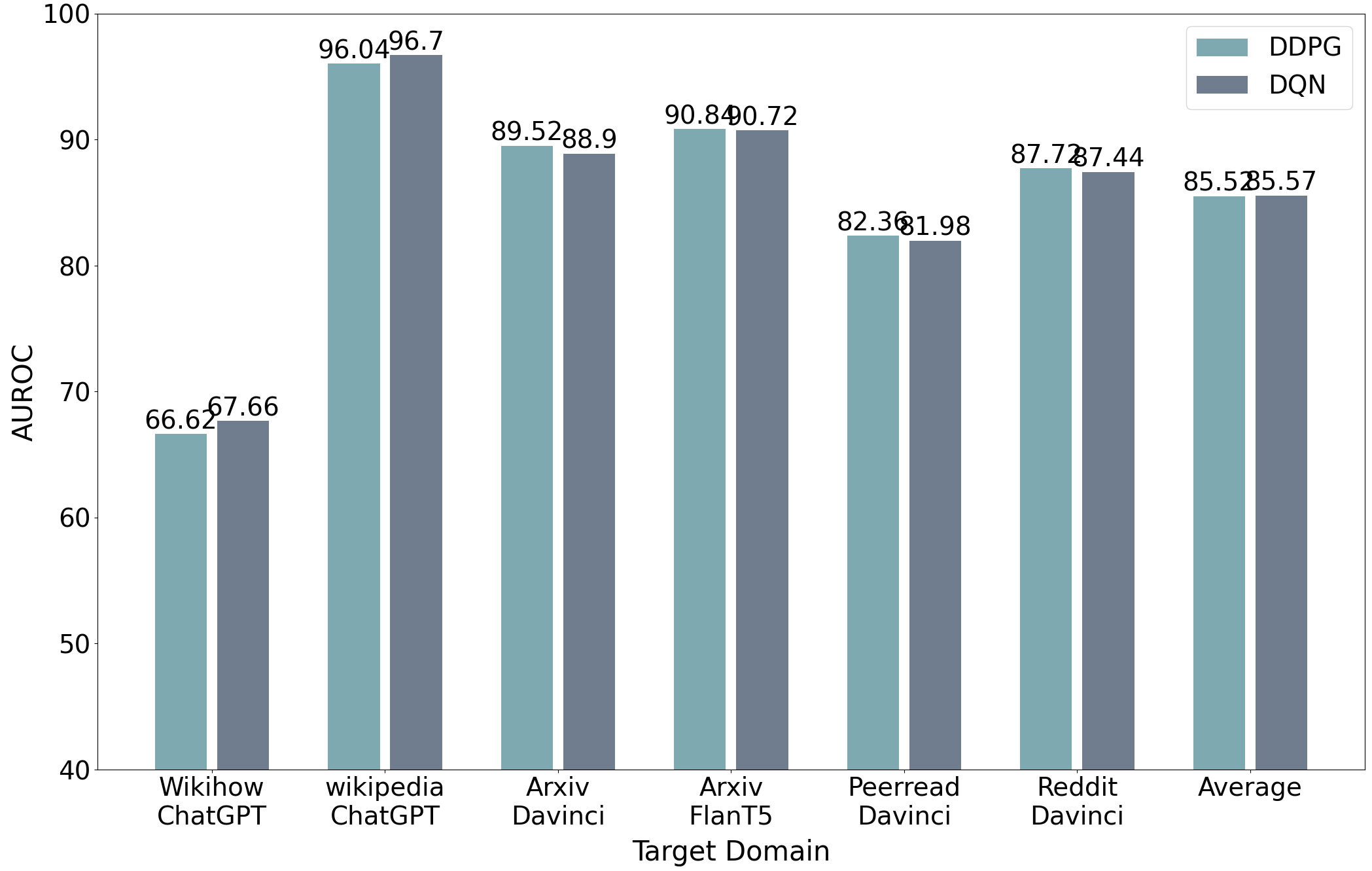}
\caption{Detection result on 6 different domains using different reinforcement learning methods to train models. The source domian is Arxic ChatGPT.}
\label{ablation_rl}
\end{figure}

\begin{figure}[t]
\centering
\includegraphics[width=1\columnwidth]{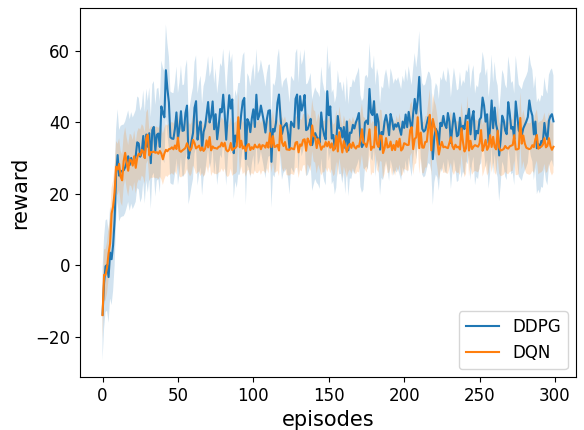}
\caption{Comparison of reward between DDPG and DQN methods.}
\label{rl}
\end{figure}

We compare DP-Net with a different reinforcement learning method, Deep Q-Network (DQN), which does not use an actor-critic framework. DQN approximates the action-value function, which estimates the potential future rewards of taking certain actions in given states. Additionally, DQN employs a target network, which is periodically updated, to provide more consistent target values during the learning process. 
As shown in the figure \ref{ablation_rl}, The models trained using the two reinforcement learning methods exhibit comparable generalization performance. Additionally, as shown in Figure \ref{rl}, the use of discrete action distributions simplifies the environment, resulting in more stable rewards over 300 iterations when using DQN compared to DDPG.

\section{Computational Complexity Analysis}
We compare the inference time of DP-Net on 2000 samples with other baselines. As shown in Table \ref{inference}, DP-Net achieves the fastest inference time, requiring only 35.53 seconds.

\begin{table}
\centering
\begin{tabular}{c|c}
\hline
 Method & inference time(s) \\
\hline
    DP-Net+G &  \textbf{35.53}\\
    Fast DetectGPT & 78.45 \\
    SCRN & 69.95\\
    RoBERTa & 37.48\\
\hline
\end{tabular}
\caption{Comparison on inference time.}
\label{inference}
\end{table}

\end{document}